\pdfoutput=1

\documentclass[11pt]{article}

\usepackage[preprint]{acl}
\usepackage{markdown}
\usepackage{times}
\usepackage{latexsym}

\usepackage[T1]{fontenc}

\usepackage[utf8]{inputenc}

\usepackage{microtype}

\usepackage{inconsolata}

\usepackage{graphicx}

\usepackage{hyperref}       
\usepackage{url}            
\usepackage{booktabs}       
\usepackage{amsfonts}       
\usepackage{nicefrac}       
\usepackage{microtype}      
\usepackage{lipsum}
\usepackage{fancyhdr}       
\usepackage{graphicx}       
\usepackage{float}
\usepackage{multirow}
\usepackage[utf8]{inputenc}
\graphicspath{{media/}}     
\usepackage{siunitx} 
\usepackage{gensymb} 
\usepackage[most]{tcolorbox}

\usepackage{markdown}
\usepackage{tabularx}
\usepackage{xcolor}
\usepackage{hyperref}
\usepackage{multirow}
\usepackage{listings}
\usepackage{longtable}
\usepackage{bbding}
\usepackage{enumitem}
\usepackage{amssymb}
\usepackage{wrapfig}
\usepackage{array}
\usepackage{diagbox}
\hypersetup{
    colorlinks=true, 
    linkcolor=blue, 
    filecolor=magenta, 
    urlcolor=cyan, 
    citecolor=green 
}

\newtcolorbox{prompt}[1]{
    enhanced,
    colback=gray!20,
    colframe=black,
    boxrule=0.3pt,
    arc=3mm,
    left=2pt,
    right=2pt,
    boxsep=3pt,
    fonttitle=\small\bfseries,
    title=#1,
    fontupper=\scriptsize
}

\definecolor{bclawblue}{RGB}{0,100,200}

\definecolor{lightgray}{gray}{0.9}
\DeclareUnicodeCharacter{FF08}{\textlparen} 
\DeclareUnicodeCharacter{FF09}{\textrparen} 

\title{AirNav: A Large-Scale UAV Vision-and-Language Navigation Dataset with Natural and Diverse Instructions}

\author{%
\parbox{\linewidth}{\centering%
\textbf{Author Name$^{1}$\thanks{Equal Contribution}} and \textbf{Author Name$^{1,2}$} \\[2ex] 
$^1$Institution Name \quad $^2$Institution Name \\[2ex] 
$^3$Institution Name \quad $^4$Institution Name \\[2ex] 
$^5$Institution Name \quad $^6$Institution Name \\[2ex] 
$^7$Institution Name \\[2ex] 
}%
}

\author{%
\parbox{\linewidth}{\centering%
\textbf{Hengxing Cai$^{1}$\thanks{Equal Contribution}}, 
\textbf{Yijie Rao$^{2}$\footnotemark[1]}, 
\textbf{Ligang Huang$^{3}$}, 
\textbf{Zanyang Zhong$^{1}$},
\textbf{Jinhan Dong$^{4}$}, \\[0.5ex]
\textbf{Jingjun Tan$^{1}$},
\textbf{Changhao Nai$^{5}$}, 
\textbf{Jue Hou$^{6}$}, 
\textbf{Wenhao Lu$^7$}
and \textbf{Renxin Zhong$^{1}$\thanks{Corresponding author}} \\[2ex]
$^1$School of Intelligent Systems Engineering, Sun Yat-Sen University \quad 
$^2$Beihang University \quad 
$^3$Peking University \quad 
$^4$Beijing University Of Posts and Telecommunications \\ 
$^5$Harbin Institute of Technology \quad 
$^6$Xiamen University \\
$^7$National University of Defense Technology \\[5ex]
}%
}

\begin{document}
\maketitle
\begin{abstract}

Existing UAV vision-and-language navigation (VLN) benchmarks rarely provide realistic aerial scenes, natural process-level instructions, and sufficient scale simultaneously, making it difficult to systematically train and evaluate UAV VLN agents under realistic settings.
To address this, we propose \textbf{AirNav}, a large-scale benchmark built on real urban aerial data, comprising 137K navigation samples with natural and diverse instructions generated via a human--LLM collaborative pipeline with 10 user personas.
We conduct a systematic evaluation of representative approaches on AirNav, ranging from traditional models to multimodal large language models (MLLMs), under unified metrics with open-source implementations.
We further propose \textbf{AirVLN-R1}, trained via supervised fine-tuning (SFT) and reinforcement fine-tuning (RFT), achieving state-of-the-art performance with a 51.82\% success rate on the test-unseen split.
Real-world experiments on a physical UAV platform provide preliminary evidence of sim-to-real transferability, and our dataset and code are publicly available.\footnote{\url{https://github.com/nopride03/AirNav}}

\end{abstract}


\section{Introduction}

Unmanned Aerial Vehicle (UAV) Vision-and-Language Navigation (VLN) aims to enable UAVs to navigate complex urban environments by following natural language instructions, supporting applications such as emergency rescue, urban patrol, and infrastructure inspection.
Compared with traditional rule-based approaches, language-guided navigation enables more flexible human--machine interaction and allows UAVs to autonomously accomplish diverse tasks across a wide range of environments.
Unlike indoor or ground-level VLN, UAV VLN requires agents to reason over long-horizon aerial views, large-scale urban landmarks, and route instructions that often include intermediate cues and conditional behaviors.
To facilitate research on this task, several benchmarks have been proposed, such as AerialVLN~\cite{liu2023aerialvlnvisionandlanguagenavigationuavs}, CityNav~\cite{lee2024citynavlanguagegoalaerialnavigation}, OpenUAV~\cite{wang2024towards}, and OpenFly~\cite{2025OpenFly}.
Recent advances in multimodal large language models (MLLMs) have opened new opportunities for language-grounded navigation; however, the lack of benchmarks that jointly provide realistic scenes, natural instructions, and sufficient scale remains a critical bottleneck for progress in UAV VLN.

\textbf{First}, realistic aerial perception requires real urban data with complex spatial structures and rich visual textures, while many existing datasets still rely on simulated environments.
\textbf{Second}, language instructions in existing datasets often lack naturalness and diversity.
Some datasets only provide target descriptions while ignoring critical intermediate cues during navigation, such as landmark references and conditional behaviors.
Others include procedural descriptions but rely heavily on template-based generation, leading to monotonous language styles that do not reflect realistic usage.
\textbf{Finally}, existing datasets are often limited in scale or language variation, hindering both the training of large-scale models and their ability to generalize across diverse environments and instruction styles.

To address these limitations, we propose \textbf{AirNav}, a large-scale UAV VLN benchmark constructed from real urban aerial data, with natural and diverse instructions produced through a human--LLM collaborative pipeline.
AirNav contains 137K navigation samples and explicitly models instruction diversity through 10 user personas.
We further conduct a unified evaluation of representative approaches, from traditional models to MLLMs, and propose \textbf{AirVLN-R1}, a navigation model optimized via a two-stage training paradigm combining supervised fine-tuning (SFT) and reinforcement fine-tuning (RFT).
AirVLN-R1 achieves state-of-the-art performance on AirNav and shows preliminary transferability on a real-world UAV platform.

The main contributions of this work are summarized as follows:

\textbf{1.} We introduce AirNav, a large-scale UAV VLN benchmark built upon real urban aerial data, featuring 137K samples with natural and diverse instructions generated via a human--LLM collaborative pipeline with 10 user personas.

\textbf{2.} A comprehensive evaluation of representative approaches---from traditional models to multimodal large language models (MLLMs)---is conducted under unified metrics, with open-source implementations released to facilitate further research.

\textbf{3.} We propose AirVLN-R1, a navigation model combining SFT and RFT, achieving state-of-the-art performance on the AirNav benchmark, including a 51.82\% success rate on the test-unseen split.

\textbf{4.} We deploy AirVLN-R1 on a physical UAV platform and conduct real-world navigation experiments across indoor and outdoor scenarios, providing initial evidence for practical deployment and sim-to-real transfer.

\section{Related Work}

Several benchmark datasets have been proposed for UAV VLN, as summarized in Table~\ref{tab:uav_dataset_comparison}.
These benchmarks exhibit different research emphases.
Some datasets are constructed from real aerial imagery, rather than simulation-based virtual environments, offering higher visual realism and greater scene diversity.
Others introduce explicit sub-goals, making them well suited for analyzing the alignment between language understanding and action decision-making.
However, these datasets typically struggle to simultaneously offer real-data-based environments, natural instructions with complete navigation processes, and sufficient data scale for comprehensive training and evaluation.
Detailed analyses are provided in Appendix~\ref{app:related_work}.




\begin{table*}[t]
\centering
\tiny
\setlength{\tabcolsep}{2pt}
\begin{tabular}{>{\centering\arraybackslash}p{2.6cm} | 
                >{\centering\arraybackslash}p{2.5cm} 
                >{\centering\arraybackslash}p{1.3cm} 
                >{\centering\arraybackslash}p{1.6cm} 
                >{\centering\arraybackslash}p{1.5cm} 
                >{\centering\arraybackslash}p{2.2cm} 
                >{\centering\arraybackslash}p{2.2cm}}
\toprule
\textbf{Dataset} & \textbf{Data Source} & \textbf{Action Space} & \textbf{Dataset Size} & \textbf{Sub-goals} & \textbf{Instruction Naturalness} & \textbf{Vocabulary Size} \\
\midrule
LANI~\cite{2018Mapping}           & Virtual                & 2 DoF & 6,000   & Yes  & Medium & 2.3K \\
AVDN~\cite{fan2022aerial}           & Real data            & 3 DoF & 3,064   & Yes & Medium & 3.3K \\
AerialVLN~\cite{liu2023aerialvlnvisionandlanguagenavigationuavs}      & Virtual                & 4 DoF & 8,446   & Yes & Medium & 4.5K \\
CityNav~\cite{lee2024citynavlanguagegoalaerialnavigation} & Real data            & 4 DoF & 32,637  & No  & N/A & 6.4K \\
OpenUAV~\cite{wang2024towards}        & Virtual                & 6 DoF & 12,149  & No & N/A & 10.8K \\
OpenFly~\cite{2025OpenFly}        & Virtual + Real data  & 4 DoF & 100k    & Yes & Medium & 15.6K \\
AirNav (Ours)  & Real data            & 4 DoF & 137k    & Yes & High & 20.3K \\
\bottomrule
\end{tabular}

\vspace{0.5em}
\scriptsize
\textbf{Note.} Here, ``Real data'' indicates that the benchmark is constructed from real aerial data, rather than synthetic or game-engine-based environments.
\caption{Comparison of representative UAV VLN benchmarks across data, scale, and instructions.}
\label{tab:uav_dataset_comparison}
\end{table*}

\section{AirNav benchmark}

\subsection{Task Definition} 
The UAV VLN task guides a UAV to complete navigation missions in environments using language instructions.
Starting from an initial position, the agent interacts with the environment over multiple steps of "Perception-Decision-Execution", during which it repeatedly predicts a series of action sequences that guide the UAV to the target.

We formulate this task as a partially observable sequential decision-making problem. 
At step $t$, the agent receives the multimodal observation:
\[
O_{\le t} = \{ v_1, \ldots, v_t;\; S_t;\; A_{1:t-1};\; L \},
\]
where $v_i$ denotes the first-person image captured by the UAV at step $i$, $S_t$ represents the current UAV state including its spatial position and heading angle, $A_{1:t-1}$ is the sequence of actions executed from the start up to the previous step, and $L$ denotes the instruction describing the target and providing path-related cues.

The agent is required to learn a policy function $\pi$ that, at each step $t$, generates an action sequence $\hat{A}_t$ conditioned on the accumulated observation $O_{\le t}$:
\[
\hat{A}_t = \pi(O_{\le t}), \quad
\hat{A}_t = \{ a_t^{(1)}, a_t^{(2)}, \dots, a_t^{(k)} \},
\]
where $a_t^{(i)}$ denotes the $i$-th action in the sequence, and $k$ is the number of actions predicted at the current step.
The model can output a variable-length sequence of discrete actions at each step, with possible actions including \textbf{move forward}, \textbf{turn left}, \textbf{turn right}, and \textbf{stop}.


\paragraph{Success Criteria.}
The task is considered successful if, after the UAV has completed the navigation, the Euclidean distance between its final position and the target location is smaller than a predefined threshold (e.g., 20 meters).


\paragraph{Metrics.}
To comprehensively evaluate navigation accuracy and path efficiency, we follow standard VLN evaluation protocols (e.g., SOON~\cite{2021SOON}, CityNav~\cite{lee2024citynavlanguagegoalaerialnavigation}) and adopt the following metrics: Navigation Error (NE), Success Rate (SR), Oracle Success Rate (OSR), and Success weighted by Path Length (SPL).
Detailed definitions are provided in Appendix~\ref{app:metrics}.

\subsection{Benchmark Construction}

\begin{figure*}[t]
  \centering
  \includegraphics[width=0.97\textwidth]{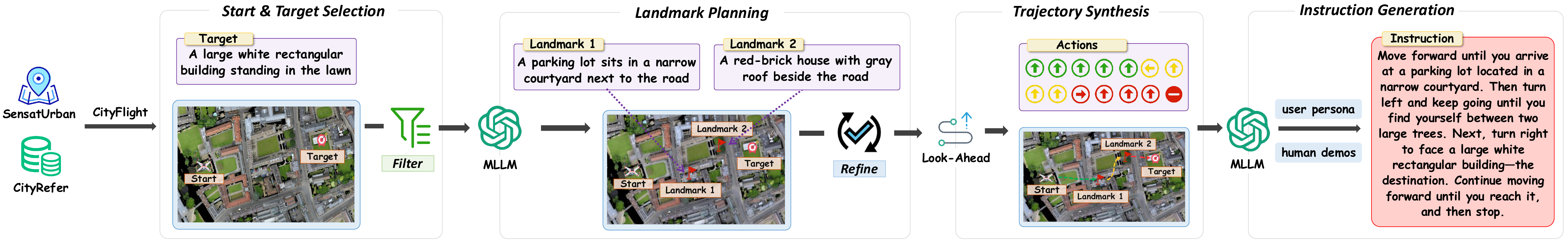} 
  \vspace{-8pt}
  \caption{Overview of the AirNav Benchmark Construction Pipeline.}
  \label{fig:data_construction}
\end{figure*}

We propose a four-step pipeline for benchmark construction, as shown in Fig.~\ref{fig:data_construction}.


\paragraph{Data Sources and Environment}
AirNav is built upon the SensatUrban~\cite{2022SensatUrban} and CityRefer~\cite{2023CityRefer} datasets. 
SensatUrban provides high-density 3D point cloud data with rich geographic structures, covering two cities, Cambridge and Birmingham. 
CityRefer supplements natural language descriptions for objects appearing in the SensatUrban.
The environment is constructed based on the CityFlight~\cite{lee2024citynavlanguagegoalaerialnavigation}, which aligns the SensatUrban data with OpenStreetMap.
CityFlight forms an interactive flight environment and provides interfaces for accessing various types of information, such as environmental images and object coordinates.
SensatUrban, CityRefer, and CityFlight are all released under the MIT License.


\paragraph{Step 1: Start and Target Selection.}
The start point is randomly sampled from the map as a feasible coordinate to initialize the navigation episode. 
A geographical object with well-defined spatial boundaries is selected as the navigation target. 
Using an MLLM, a natural language description of the target is generated from the endpoint's perspective, and samples with ambiguous or confusing target descriptions are filtered out.

\paragraph{Step 2: Landmark Planning.}
Given the selected start and target locations, the MLLM is prompted to identify representative geographic objects between them and generate corresponding descriptions as intermediate landmarks.
To preserve perceptual continuity along the route, we impose a maximum distance constraint between consecutive landmarks and discard samples in which landmarks are overly sparse.
In addition, we perform semantic refinement for landmarks. 
For each landmark, the model further verifies the factual correctness of its description and rewrites it to ensure semantic clarity and disambiguation.

\paragraph{Step 3: Trajectory Synthesis.}
For each pair of consecutive nodes, such as the start point and the first landmark or two adjacent landmarks, we apply a look-ahead strategy~\cite{liu2023aerialvlnvisionandlanguagenavigationuavs} to generate an executable action sequence for the corresponding path segment. 
All segment-level action sequences are then concatenated to form a complete trajectory that spans from the start to the target.

\paragraph{Step 4: Instruction Generation.}
The trajectory, map, and the spatial positions and semantic descriptions of all nodes are provided as inputs to GPT-4o, which generates navigation instructions covering target descriptions, path guidance, spatial relations, and trigger conditions. 
To model linguistic realism and diversity, following principles from User-Centered Design~\cite{Norman2013The,2003Personas} and sociolinguistic studies on linguistic variation~\cite{labov1973sociolinguistic,tagliamonte2011variationist}, we construct 10 representative user personas (see Appendix~\ref{app:personas}) based on age group, social role, and expression preferences, covering typical urban navigation scenarios and diverse language styles.
During instruction generation, persona-specific settings are incorporated to guide GPT-4o to produce navigation instructions with diverse language styles and variations in expression.
To further improve linguistic naturalness, human-authored real navigation instructions are included as few-shot examples within the prompt.

\paragraph{Quality Control.}
To ensure data quality at scale, we adopt an iterative quality control process throughout the pipeline.
After each round of generation, we randomly sample examples and manually inspect outputs across all stages, including the validity of start and target points, the clarity of target descriptions, the spatial consistency of landmark annotations, and the continuity of synthesized trajectories.
Identified issues are categorized by error type (e.g., ambiguous landmarks, visually indistinguishable targets, excessive gaps between consecutive landmarks), and the corresponding generation prompts or filtering rules are updated accordingly.
The data is then regenerated under the revised settings.
This inspect-revise cycle is repeated over multiple iterations until the sampled error rate stabilizes at a low level, after which we proceed to large-scale generation.

\subsection{Dataset and Instruction Analysis} 

\begin{figure*}[t]
  \centering
  \includegraphics[width=0.94\textwidth]{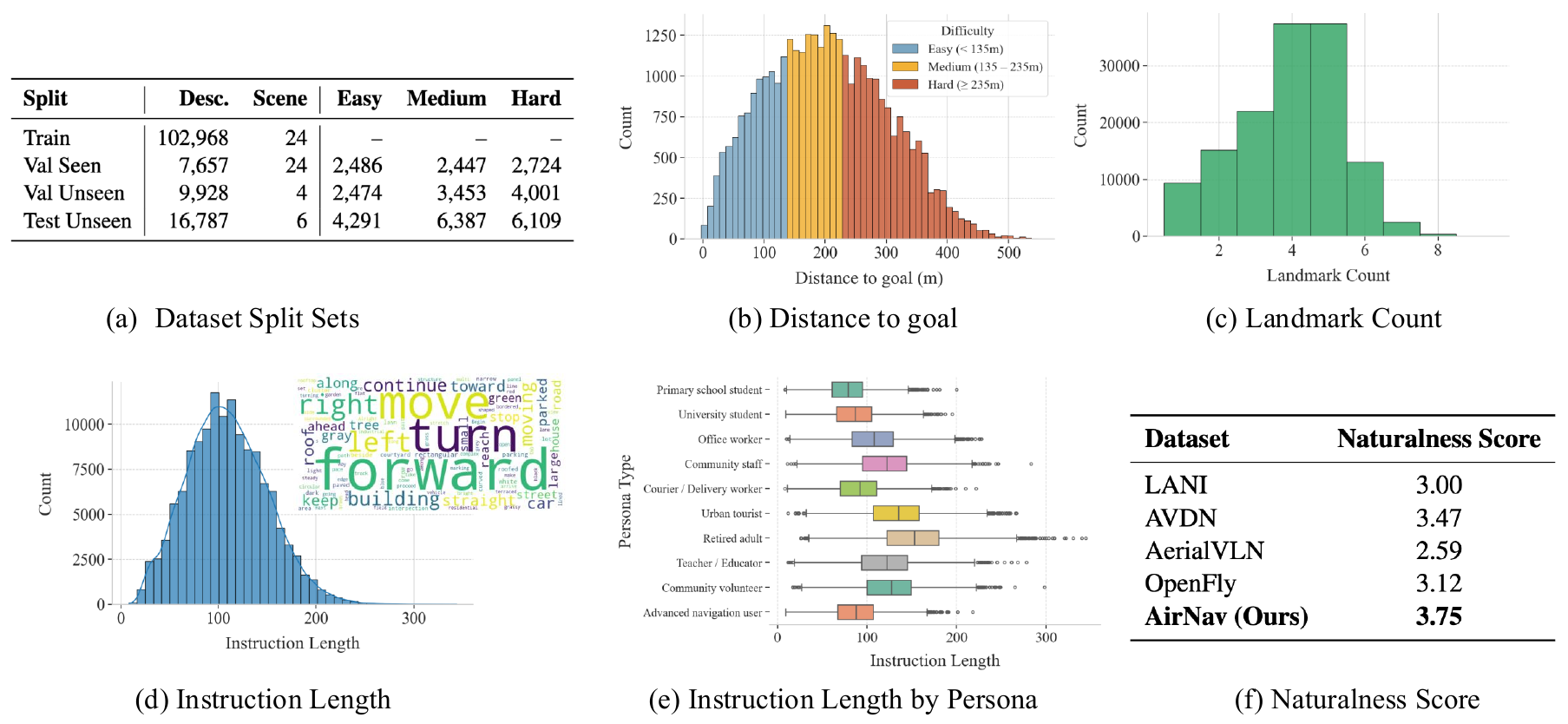} 
  \vspace{-8pt}
  \caption{Dataset Analysis and Instruction Naturalness of AirNav.}
  \label{fig:data_analysis}
\end{figure*}

\paragraph{Dataset Splits.}
Following evaluation protocols in prior UAV VLN benchmarks~\cite{liu2023aerialvlnvisionandlanguagenavigationuavs, lee2024citynavlanguagegoalaerialnavigation}, we split the AirNav dataset into four subsets: Train, Validation Seen (val-seen), Validation Unseen (val-unseen), and Test Unseen (test-unseen). 
The val-seen split shares the same environments as Train and is used to evaluate model performance in known scenes, while val-unseen and test-unseen are sampled from novel environments to assess generalization under unseen conditions.
Fig.~\ref{fig:data_analysis} (a) summarizes key statistics for each subset.


\paragraph{Distance Distribution and Task Difficulty.}
We categorize navigation paths into three difficulty levels based on path length percentiles of the training set: \textit{Easy} ($<135$m), \textit{Medium} (135--230m), and \textit{Hard} ($\geq230$m).
Fig.~\ref{fig:data_analysis} (b) presents the distance distributions for different difficulty categories over the entire evaluation set.

\paragraph{Number of Intermediate Landmarks.}
As shown in Fig.~\ref{fig:data_analysis} (c), most trajectories contain between 2 and 6 intermediate landmarks, with 4 to 5 being the most prevalent, supporting step-wise modeling of the navigation process and spatial-semantic alignment between instructions and actions.


\paragraph{Instruction Length and Vocabulary Statistics.}
As shown in Fig.~\ref{fig:data_analysis} (d), instruction lengths span a wide range, with a peak around 100 words. 
The dataset includes both concise instructions with compact structures and longer instructions that provide detailed descriptions of intermediate operations, reflecting the coexistence of different information densities and narrative granularities. 
In addition, AirNav exhibits a vocabulary size of 20.3k, which is significantly larger than that of existing UAV VLN datasets, indicating higher linguistic diversity and reduced repetition in instruction expressions.

\paragraph{Persona-conditioned Instruction Characteristics.}

AirNav explicitly models instruction diversity through the introduction of user personas. 
This design captures systematic differences among user groups in both instruction length and information organization, providing a more comprehensive representation of human instruction behaviors across varying information densities and narrative granularities. 
As illustrated in Figure~\ref{fig:data_analysis} (e), distinct personas show clear separation in the median values and distribution ranges of instruction length. 
For example, Retired Adults tend to produce longer and more explanatory instructions, whereas University Students or Advanced Navigation Users generally prefer concise expressions. 
In Appendix~\ref{app:case_study_persona_specific_instructions}, we further conduct a case-study analysis of instructions generated under different personas.

\paragraph{Instruction Naturalness Analysis.}
To quantitatively evaluate instruction naturalness, we conduct an LLM-based automated assessment, sampling 2,000 instructions per dataset and scoring them on a 5-point scale using GPT-4o; evaluation details and the scoring prompt are provided in Appendix~\ref{app:prompt_naturalness_evaluation}.
To validate the reliability of this evaluation, we further conduct a human annotation study, achieving a Krippendorff's $\alpha$ of 0.70 (inter-annotator agreement) and a Spearman's $\rho$ of 0.74 (human--LLM correlation), confirming strong consistency with human judgments (see Appendix~\ref{app:human_annotation}).
As shown in Figure~\ref{fig:data_analysis} (f), AirNav achieves the highest naturalness score (3.75), significantly outperforming all other benchmarks, indicating that its instructions more closely resemble natural language requests from real users rather than templated formulations.
A case-study comparison of instructions from different datasets is provided in Appendix~\ref{app:case_study_instruction_naturalness}.


\section{AirVLN-R1 Model}


\subsection{Overall Architecture}
\begin{figure*}[t]
  \centering
  \includegraphics[width=0.95\textwidth]{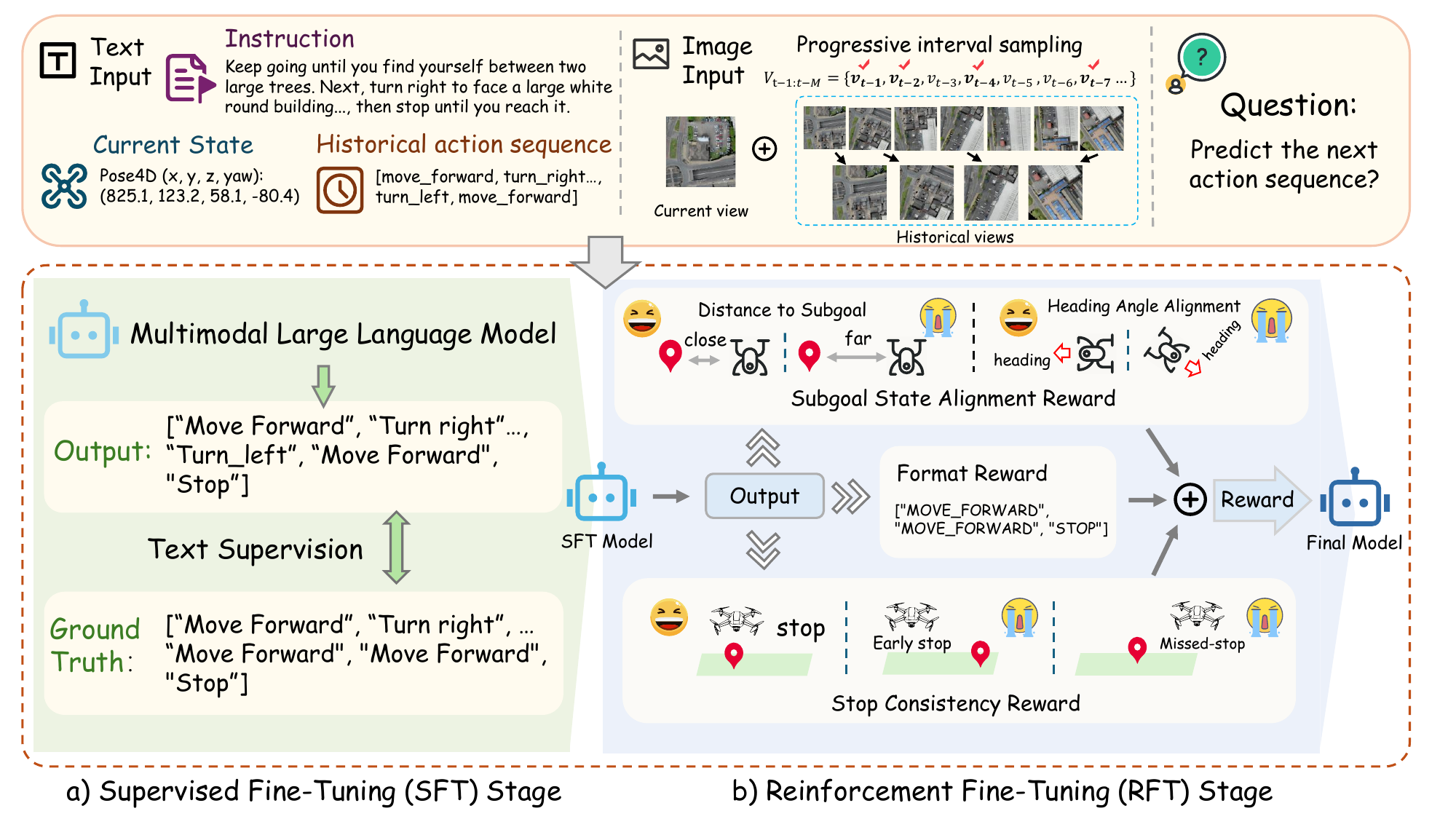} 
  \vspace{-8pt}
  \caption{Overview of the AirVLN-R1 architecture. The model receives multimodal inputs and predicts an action sequence to control the UAV. A two-stage training paradigm is used to enhance performance.}
  \label{fig:AirVLN_R1}
\end{figure*}

We model the UAV VLN task as a multi-step "Perception-Decision-Execution" loop. 
As shown in Fig.~\ref{fig:AirVLN_R1}, at each step, the model receives multimodal input and predicts a sequence of actions to control the UAV's movement, continuing until the output is \textbf{stop} or the maximum number of steps is reached.
To further optimize the policy, we adopt a dedicated two-stage training strategy.

\subsection{Input, Output and Prompt Design}
\label{sec:input_output_and_prompt}

\subsubsection{Input}
At each step, the AirVLN-R1 model processes multimodal inputs, which include both textual and visual information.

For the textual input, it consists of the following three components:
(1) \textbf{Instruction}: A natural language description of the target location and path clues;
(2) \textbf{Current State}: The UAV's coordinates and heading angle;
(3) \textbf{Historical Action Sequence}: The sequence of actions already executed by the UAV from the starting point to the current step.

The visual input provides the perception of the surrounding environment, including:
(1) \textbf{Current View Image}: An image captured from the UAV's first-person perspective at the current step;
(2) \textbf{Historical View Images}: A set of key images selected from historical observations to construct a visual memory. 
To control the input size and enhance information efficiency, we adopt a \textbf{Progressive Interval Sampling} strategy for visual memory construction.
See Section~\ref{sec:history-frame-selection} for details.

\subsubsection{Output}
The model outputs a variable-length sequence of up to 8 discrete actions, representing the actions required to proceed from the current state.

\subsubsection{Prompt Design}

We design a structured prompt template specifying the task role, input structure, output requirements, and action space definition, with the complete template provided in Appendix~\ref{app:prompt}.

\subsection{Historical View Image Selection}
\label{sec:history-frame-selection}
Directly incorporating the full sequence of historical view images introduces substantial input redundancy and computational overhead. 
To address this issue, we propose a lightweight historical view image sampling mechanism---\textbf{Progressive Interval Sampling}, 
which preserves dense observations from recent steps while sparsely sampling distant ones, thereby reducing the historical input size 
without losing critical contextual information.

At step $t$, we select at most $N$ historical view images from the past observations.
Sampling starts from the most recent view and proceeds backward with an interval that grows linearly over time. 
Specifically, the offset of the $i$-th selected historical view is defined recursively as:
\[
s_i = s_{i-1} + i, \quad s_{-1} = 1,\quad i = 0, 1, \dots, N-1.
\]

Based on the offsets $\{s_i\}$, the visual memory at step $t$ is constructed as:
\[
\mathcal{H}_t = \{ v_{t - s_i} \mid 0 \le i \le N-1,\; t - s_i \ge 1 \},
\]
where $v_{t - s_i}$ represents the view image captured at step $t - s_i$.





\subsection{Training Paradigm}

To enhance the model's perceptual understanding and decision--making capability, AirVLN-R1 adopts a two-stage training paradigm inspired by DeepSeek-R1~\cite{deepseekai2025deepseekr1incentivizingreasoningcapability}, consisting of SFT followed by RFT.

\subsubsection{Supervised Fine-Tuning}
In the first stage, we fine-tune the model end-to-end on the training set following the prompt format defined in Section~\ref{sec:input_output_and_prompt}, guiding the model to build a mapping from multimodal inputs to executable action sequences via next-token prediction with cross-entropy loss.



\subsubsection{Reinforcement Fine-Tuning}
From a decision-making perspective, UAV VLN requires the agent to 
(i) make consistent progress toward intermediate subgoals, 
(ii) reliably determine when to terminate the episode, and 
(iii) generate valid outputs.
To explicitly capture these requirements, we design a multi-objective reward function, where each reward component corresponds to one of the above decision objectives.
We use Group Relative Policy Optimization (GRPO) to optimize the policy.


\paragraph{1. Subgoal State Alignment Reward}
The UAV state at each intermediate subgoal of the trajectory represents a reasonable position and heading angle achieved by executing a sequence of expert actions.
We aim for the model to approach these subgoal states through its own predicted action sequences. 
To this end, we design two rewards reflecting position proximity and heading angle alignment.

\subparagraph{(1) Distance-to-Subgoal}
This reward encourages the UAV to reduce its distance to the subgoal by executing the predicted action sequence at step $t$, resulting in a closer state at step $t+1$.
The reward is defined as:
\[
r_{\text{dis}} = \max\left( \frac{d_{t} - d_{t+1}}{d_{t} + \epsilon},\, 0 \right),
\]
where $d_t$ denotes the Euclidean distance between the UAV and the subgoal before executing the predicted action sequence at step $t$,
and $d_{t+1}$ denotes the distance after the execution.

\subparagraph{(2) Heading Angle Alignment}


This reward measures whether the UAV's heading angle after executing the action sequence is aligned with the subgoal's heading angle. 
Let $\Delta_{\text{yaw}}$ denote the angular difference between the
two headings, normalized to $(-180^\circ, 180^\circ]$. 
The reward is defined as:
\[
r_{\text{yaw}} = \max\left( 1 - \frac{|\Delta_{\text{yaw}}|}{\tau_{\text{yaw}}},\, 0 \right),
\]
where $\tau_{\text{yaw}}$ controls the tolerance range (e.g., $60^\circ$).



\paragraph{2. Stop Consistency Reward.}
An incorrect \textbf{stop} decision---whether premature or missed---directly causes task failure.
To mitigate such errors, we define the following reward mechanism: the reward is \(\alpha\) if both the predicted and ground-truth action sequences end with \textbf{stop}; the reward is \(\beta\) if neither the predicted nor the ground-truth sequence ends with \textbf{stop}; otherwise, the reward is 0.




\paragraph{3. Format Reward.}


To encourage the model to produce outputs with correct structure and valid syntax, we assign a constant reward $r_{\text{format}}=\gamma$ if the output is well-formed; otherwise, we set $r_{\text{format}}=0$.

\paragraph{Overall Reward Function Definition.}
The final overall reward is obtained by linearly combining all the aforementioned reward components:
\[
r_{all} = \lambda_1 \cdot r_{\text{dis}} 
    + \lambda_2 \cdot r_{\text{yaw}} 
    + r_{\text{stop}} 
    + r_{\text{format}}.
\]
\section{Experiments}

\subsection{Experimental Settings}
\subsubsection{Baseline Models}
We select several representative baseline models, covering traditional methods, zero-shot MLLMs, and recent navigation models.
Detailed information about all baseline models can be found in Appendix~\ref{app:baselines}.

\subsubsection{Implementation Details}
AirVLN-R1 is built upon Qwen2.5-VL-7B and trained on an 8×A100 GPU server. The hyperparameter settings for both training stages are provided in Appendix~\ref{app:hyperparameter_settings}.
To ensure fair comparison, all MLLM baselines are provided with the same inputs as AirVLN-R1, including the navigation instruction, current UAV state, historical action sequence, current view image, and historical view images selected by the same progressive interval sampling strategy, as well as the same prompts.
The temperature settings and other hyperparameters for all MLLM baselines are available in our open-source repository.



\subsection{Main Results}


The comparative results on AirNav are shown in Table~\ref{tab:simulation_result}.
Traditional methods perform poorly, with Seq2Seq and CMA achieving SR below 2\% and 6\% respectively, failing to handle the complexity of UAV VLN.
Among zero-shot MLLMs, LLaMA-3.2-11B-Vision yields similarly low SR, while GPT-4o performs better but remains limited on unseen scenarios.
The Qwen series achieves stronger results with SR scaling consistently with model size, and Qwen3-VL-235B-A22B reaches the highest SR among all zero-shot baselines.

$\pi_0$, fine-tuned on AirNav data, achieves only limited performance (SR of 6.29\% on val-seen), suggesting that Vision-Language-Action (VLA) models designed for robot manipulation scenarios do not transfer well to UAV VLN.
Uni-NaVid, designed specifically for embodied navigation and fine-tuned on AirNav data, reaches an SR of 20.56\% on val-seen and 15.89\% on test-unseen, reflecting the advantage of navigation-oriented model design.

AirVLN-R1 achieves the best performance across all metrics in the AirNav Benchmark, with an SR of 51.82\% on test-unseen.
Notably, AirVLN-R1, built upon Qwen2.5-VL-7B, achieves superior performance compared to larger-scale models such as Qwen2.5-VL-32B and Qwen3-VL-235B-A22B, underscoring the effectiveness of task-specific supervision and RFT.


\begin{table*}[t]
\centering
\tiny
\setlength{\tabcolsep}{2pt}
\begin{tabular}{>{\centering\arraybackslash}p{3.4cm} | >{\centering\arraybackslash}p{0.77cm} >{\centering\arraybackslash}p{0.77cm} >{\centering\arraybackslash}p{0.77cm} >{\centering\arraybackslash}p{0.77cm} | >{\centering\arraybackslash}p{0.77cm} >{\centering\arraybackslash}p{0.77cm} >{\centering\arraybackslash}p{0.77cm} >{\centering\arraybackslash}p{0.77cm} | >{\centering\arraybackslash}p{0.77cm} >{\centering\arraybackslash}p{0.77cm} >{\centering\arraybackslash}p{0.77cm} >{\centering\arraybackslash}p{0.77cm}}
\toprule
\textbf{Method} & \multicolumn{4}{c|}{\textbf{Validation Seen}} & \multicolumn{4}{c|}{\textbf{Validation Unseen}} & \multicolumn{4}{c}{\textbf{Test Unseen}} \\
& NE$\downarrow$ & SR$\uparrow$ & OSR$\uparrow$ & SPL$\uparrow$ & NE$\downarrow$ & SR$\uparrow$ & OSR$\uparrow$ & SPL$\uparrow$ & NE$\downarrow$ & SR$\uparrow$ & OSR$\uparrow$ & SPL$\uparrow$ \\
\midrule
Random                    & 222.0 & 0.76  & 5.68  & 0.67  & 223.1 & 0.76  & 4.73  & 0.68  & 217.2 & 0.77  & 5.57  & 0.69  \\
Seq2Seq$^*$                   & 321.2 & 1.65  & 9.48  & 1.44  & 343.5 & 0.92  & 9.92  & 0.73  & 334.5 & 1.32  & 10.69 & 1.13  \\
CMA$^*$                       & 184.7 & 5.21  & 15.84 & 4.80  & 201.1 & 3.88  & 16.07 & 3.47  & 189.4 & 4.63  & 17.48 & 4.16  \\
\midrule
Qwen2.5-VL-7B             & 182.1 & 1.96  & 2.36  & 1.82  & 192.5 & 1.59  & 1.76  & 1.41  & 184.2 & 1.70  & 1.95  & 1.53  \\
Qwen2.5-VL-32B            & 160.9 & 3.25  & 3.66  & 2.94  & 170.6 & 2.68  & 2.98  & 2.38  & 162.4 & 2.99  & 3.25  & 2.68  \\
Qwen3-VL-235B-A22B        & 157.6 & 5.58  & 9.23  & 5.18  & 168.5 & 5.14  & 8.42  & 4.63  & 155.6 & 5.21  & 8.44  & 4.75  \\
LLaMA-3.2-11B-Vision      & 182.3 & 1.02  & 4.85  & 0.95  & 193.8 & 1.57  & 4.51  & 1.41  & 185.5 & 1.31  & 4.28  & 1.12  \\
GPT-4o                    & 154.0 & 4.71  & 8.68  & 4.22  & 164.3 & 4.16  & 7.18  & 3.73  & 156.2 & 4.56  & 7.88  & 4.13  \\

\midrule
$\pi_0$$^*$ & 144.2 & 6.29 & 14.27 & 5.40 & 163.6 & 4.47 & 13.37 & 3.90 & 147.5 & 5.28 & 13.93 & 4.60 \\
Uni-NaVid$^*$ & 112.3 & 20.56 & 38.02 & 18.45 & 129.3 & 16.60 & 33.45 & 14.71 & 131.2 & 15.89 & 34.25 & 14.18 \\
Qwen2.5-VL-7B SFT-only$^*$   & 45.4  & 44.27 & 54.96 & 43.04 & 48.7  & 40.98 & 52.57 & 39.89 & 47.5  & 40.20 & 53.21 & 39.15 \\
Qwen2.5-VL-7B RFT-only$^*$   & 165.1 & 2.38  & 4.91  & 2.14  & 173.2 & 2.06  & 3.90  & 1.82  & 164.0 & 2.41  & 4.57  & 2.13  \\
\midrule
AirVLN-R1$^*$ (Ours)         & \textbf{39.7} & \textbf{51.76} & \textbf{61.45} & \textbf{50.61} & \textbf{40.8} & \textbf{51.32} & \textbf{61.78} & \textbf{50.11} & \textbf{39.8} & \textbf{51.82} & \textbf{62.73} & \textbf{50.66} \\

\bottomrule
\end{tabular}
\caption{Comparison of model performance across evaluation scenarios. $^*$ denotes models fine-tuned on AirNav.}
\label{tab:simulation_result}
\end{table*}

\subsection{Persona-Level Analysis}

We evaluate the SR of all models across all 10 user personas on the test-unseen split; complete results are provided in Appendix~\ref{app:persona_results}.
We observe that persona is associated with differences in SR.
Among all personas, Teacher/Educator and Advanced Navigation User instructions yield the top two SR values: the former benefit from well-structured, step-by-step organization that facilitates progressive subgoal alignment, while the latter are concise and goal-oriented with minimal linguistic noise, making key navigation cues easy to extract.
Conversely, Retired Adult instructions tend to be verbose with repeated contextual descriptions that increase the difficulty of extracting critical navigation cues, while Primary School Student instructions, though concise, lack sufficient spatial detail; both personas consequently achieve comparatively lower SR.

\subsection{Ablation Study}

\paragraph{Training Paradigm.}
We compare three paradigms: SFT-only, RFT-only, and SFT+RFT, as shown in Table~\ref{tab:simulation_result}.
SFT-only achieves competitive performance on val-seen (SR = 44.27\%), but leaves room for further improvement on val-unseen and test-unseen splits, indicating that trajectory imitation alone does not fully generalize.
RFT-only struggles due to highly sparse reward signals without SFT initialization, converging to suboptimal policies.
SFT+RFT achieves the best and most stable results across all splits, demonstrating that the two stages are complementary.

\paragraph{Historical View Image Selection.}
We compare four strategies for incorporating historical observations: No-History, Last-K, Uniform-K, and our Progressive Interval Sampling.
Progressive Interval Sampling achieves the best performance (SR = 51.82\% on test-unseen), outperforming Uniform-K (49.83\%) by better balancing fine-grained short-term perception with long-range contextual modeling.
Full results are provided in Appendix~\ref{app:ablation_history}.

\paragraph{Reward Component Analysis.}
We ablate each component of the RFT reward function individually.
Removing the Subgoal State Alignment reward causes the most significant degradation (SR: 51.82\% $\rightarrow$ 46.00\%), confirming its central role in guiding path planning.
Removing the Stop Consistency reward also leads to a notable drop (SR: 51.82\% $\rightarrow$ 47.65\%), highlighting the importance of explicit termination supervision.
The Format reward contributes to training stability with a smaller but consistent effect.
Detailed results are in Appendix~\ref{app:ablation_reward}.

\paragraph{Sensitivity Analysis of Reward Weights.}
We conduct a $3{\times}3$ grid search over the key weighting parameters $\lambda_1$ and $\lambda_2$.
Performance remains stable across all tested configurations, with a maximum SR fluctuation of within 1.5 percentage points, indicating that the method is robust to the precise choice of reward weights.
See Appendix~\ref{app:ablation_sensitivity} for the full sensitivity results.

\subsection{Real World Experiments}

\subsubsection{Setup}
To evaluate the performance and deployment capability of AirVLN-R1 in real-world environments, we conduct a series of physical tests without any additional fine-tuning.
The experiments cover two typical settings---indoor and outdoor---each containing 100 navigation tasks with varying levels of path complexity.
Examples of real-world tasks, along with the details of the experimental setup, are provided in Appendix~\ref{app:real_world_setup_and_cases}.


\subsubsection{Results}


\paragraph{Quantitative results and overall comparison.}
We evaluate different methods in real-world test, with detailed results provided in Appendix~\ref{app:real_world_results}.
Traditional methods nearly fail to complete any real-world tasks, with Seq2Seq and CMA succeeding on only 1 and 2 out of 200 tasks respectively.
In contrast, general MLLMs under zero-shot settings can only solve a very small number of tasks.
The stronger model GPT-4o shows a noticeable improvement, yet its overall SR remains limited.
Compared with all baselines, AirVLN-R1 reaches SR$=53/200$ and achieves the lowest NE (70.6), maintaining a consistent relative advantage in the real-world environment that aligns with the simulation evaluation.




\paragraph{Resource Cost and Inference Efficiency.}
Traditional baselines exhibit low computational overhead but completely fail in real-world navigation, while large-scale MLLMs incur substantial latency and hardware requirements, making real-time deployment challenging.
In contrast, AirVLN-R1 achieves the best real-world SR with moderate inference latency and GPU memory consumption, representing a more balanced and deployable solution for UAV VLN.
A detailed comparison of computational cost and efficiency is provided in Appendix~\ref{app:resource_cost}.



\section{Conclusion}
This paper presents AirNav, a large-scale UAV VLN benchmark constructed from real urban aerial data, featuring 137K navigation samples with natural and diverse language instructions, along with a comprehensive evaluation of representative approaches under unified metrics and open-source implementations.
We further propose AirVLN-R1, a navigation model combining SFT and RFT that achieves state-of-the-art performance on the AirNav benchmark.
Real-world experiments validate the feasibility and consistency of sim-to-real transfer on physical UAV platforms.
We hope AirNav provides a realistic and comprehensive evaluation foundation for future research on UAV VLN.

\newpage
\section{Limitations}

\subsection{Limited Data Sources and Scenario Coverage}
AirNav is primarily constructed based on SensatUrban and CityRefer. While these data sources provide high-fidelity urban scenes, their geographic coverage, urban styles, and infrastructure layouts are inherently constrained by the regions and annotation schemes of the existing datasets. As a result, the generalization ability of models across different cities, countries, and seasonal conditions remains to be validated with more diverse data. In addition, aerial imagery often exhibits relatively fixed viewpoints and altitude distributions, which may not fully capture the more complex altitude variations, occlusion patterns, and extreme lighting conditions encountered in real-world UAV VLN tasks.

\subsection{Gap Between Discrete Action Modeling and Real Flight Control}
We adopt a discrete action set and allow the model to output an action sequence of up to eight steps at each step. This design facilitates stable training and fair comparison across methods, but it cannot fully represent the fine-grained motions and dynamic constraints involved in real UAV continuous control. In long-horizon navigation or tasks requiring precise target approach, discrete actions may introduce trajectory approximation errors, thereby limiting the policy’s performance in complex maneuvering scenarios.

\subsection{Gap Between Simulation Evaluation and Real-World Deployment}
Although we conduct preliminary real-world flight experiments across 200 navigation tasks, the environmental diversity of the real-world evaluations remains limited. Moreover, differences between simulation and real UAV deployment persist in terms of perception noise, viewpoint variations, and control uncertainty. Consequently, performance improvements observed on the benchmark do not necessarily translate directly into stable gains in real-world settings. Further reducing the sim-to-real gap remains an important direction for future work.

\subsection{Reproducibility of Closed-Source Model Results}
Our evaluation includes closed-source models such as GPT-4o. As these models are subject to API access restrictions and potential version updates over time, exact reproduction of the reported results may be limited.
\section{Ethical Considerations}
\subsection{Privacy Risks in Urban Aerial Data}
AirNav relies on real-world urban scene data, and aerial imagery may implicitly contain sensitive areas or information related to human activities. Although this work uses datasets intended for research purposes, the release of the benchmark and models should be handled with care to avoid unintended use in unauthorized area localization, surveillance, or privacy inference.

\subsection{Reliability in Safety-Critical Applications}
UAV VLN is a typical safety-critical application. Even when a model performs well on benchmark evaluations, real-world deployment may still suffer from perception errors or instruction misinterpretation, potentially leading to collisions or entry into hazardous areas. Therefore, the methods proposed in this work should not be regarded as a complete system that can directly replace human control, but rather as components that need to be used in conjunction with engineering-level safety mechanisms, such as human intervention and geofencing.

\subsection{Dual-Use Concerns and Potential Misuse}
UAV VLN technology exhibits clear dual-use characteristics. While it can support positive applications such as search-and-rescue and infrastructure inspection, it may also be misused for surveillance, tracking, or other inappropriate purposes. We emphasize that the benchmark and models are intended solely for research and lawful applications, and we recommend clearly specifying usage scope and restrictions when releasing code and models to mitigate potential misuse risks.

\newpage
\bibliography{custom}

\newpage
\newpage
\appendix

\clearpage
\section{Related Work}
\label{app:related_work}
This section provides a detailed analysis of representative UAV VLN benchmarks, summarizing their design characteristics as well as their strengths and limitations.

LANI~\cite{2018Mapping} is one of the earliest UAV VLN benchmarks. 
It is constructed in an open simulated environment and supports the evaluation of basic path-following capabilities.
However, due to its simplified scenes and the lack of explicit spatial structure and photorealism, the overall task difficulty is relatively low, making it insufficient to reflect the complexity of real-world navigation scenarios.

AVDN~\cite{fan2022aerial} introduces a multi-turn natural language interaction mechanism that emphasizes human-robot collaboration. 
The dataset is collected in real-world environments, which enhances the realism of language interactions. 
Nevertheless, AVDN suffers from limited data scale and relatively short navigation paths. 
Moreover, its reliance on manually designed dialogue procedures makes it difficult to directly apply to large-scale evaluation settings.

AerialVLN~\cite{liu2023aerialvlnvisionandlanguagenavigationuavs} is constructed using multiple simulated environments, and its instruction design explicitly includes intermediate navigation steps. 
This formulation facilitates the study of alignment between language and actions. However, as the dataset is entirely based on simulation, it lacks the visual details of real urban environments, making it challenging to evaluate model generalization performance under practical conditions.

CityNav~\cite{lee2024citynavlanguagegoalaerialnavigation} is a large-scale dataset sourced from real aerial imagery of urban environments, offering strong realism in terms of visual texture and scene diversity. However, its instructions mainly focus on target descriptions and do not include intermediate navigation processes, which limits its effectiveness in evaluating models' spatial reasoning and step-by-step navigation capabilities.

OpenUAV~\cite{wang2024towards} is built on a high-fidelity simulation platform that supports 6-DoF flight and multi-view perception. 
Its instructions are generated by LLMs and subsequently refined by human annotators, resulting in task settings that are closer to real navigation requirements. Nonetheless, the dataset remains dependent on virtual environments, and its overall scale is relatively limited, restricting coverage of complex and diverse real scenarios.

OpenFly~\cite{2025OpenFly} leverages multi-source rendering engines to achieve high visual diversity and realism, and employs an automated toolchain to generate large-scale navigation trajectories and instructions, significantly improving data construction efficiency and task complexity. 
However, its instructions are entirely model-generated, lacking the linguistic habits and stylistic variations of human language. 
As a result, the overall naturalness and diversity of instruction expressions remain limited.

\section{Evaluation Metrics}
\label{app:metrics}

This section provides detailed definitions of the evaluation metrics used in our experiments.
\begin{itemize}
    \item \textbf{Navigation Error (NE)}: the Euclidean distance between the agent's final position and the ground-truth target location. Lower values indicate higher accuracy.
    \item \textbf{Success Rate (SR)}: the proportion of episodes in which the agent terminates within the target distance threshold.
    \item \textbf{Oracle Success Rate (OSR)}: the proportion of trajectories that enter the target threshold at any step, regardless of whether the agent stops correctly.
    \item \textbf{Success weighted by Path Length (SPL)}: a success metric that penalizes redundant paths, encouraging shorter and more efficient trajectories.
\end{itemize}

\section{User Personas for Navigation Instruction Generation}
\label{app:personas}

The user personas and their corresponding language style preferences are summarized in Table~\ref{tab:user_personas}.

\begin{table*}[t]
\centering
\resizebox{\textwidth}{!}{
\begin{tabular}{c|l|c|l|l}
\hline
\textbf{ID} & \textbf{Persona} & \textbf{Age Group} & \textbf{Social Role / Background} & \textbf{Preferred Language Style} \\
\hline
P1 & Primary school student & Child & Student with limited spatial experience & Simple wording, action-oriented, highly explicit instructions \\
P2 & University student & Young adult & Student, frequent user of map-based tools & Concise and direct, efficiency-focused, path-oriented terminology \\
P3 & Office worker & Young to middle-aged & Daily commuter, familiar with urban structure & Structured expression, preference for optimal routes \\
P4 & Community staff & Middle-aged & Property management / security / administrative staff & Stable phrasing, safety reminders, familiarity with local landmarks \\
P5 & Courier / Delivery worker & Young to middle-aged & High-frequency navigation user & Highly efficient instructions, emphasis on path clarity and ordering \\
P6 & Urban tourist & Young to middle-aged & Temporary visitor, unfamiliar with the environment & Landmark-rich descriptions, strong explanations, contextual guidance \\
P7 & Retired adult & Elderly & Non-technical user, slower interaction pace & Redundant explanations, repeated reminders, focus on safety and comfort \\
P8 & Teacher / Educator & Middle-aged & Emphasis on logic and clarity & Formal language, well-structured instructions, pedagogical tone \\
P9 & Community volunteer & Middle-aged to elderly & Active resident, frequent participation in public affairs & Friendly phrasing, landmark-oriented guidance, everyday language \\
P10 & Advanced navigation user & Young to middle-aged & Navigation expert / experienced user & Minimal technical language, clear structure, preference for optimal routes \\
\hline
\end{tabular}
}
\caption{User personas and language style preferences for navigation instruction generation.}
\label{tab:user_personas}
\end{table*}

\section{Case Study on Persona-Specific Instructions}
\label{app:case_study_persona_specific_instructions}

We further conduct a qualitative analysis through case studies to examine differences in instructions generated under different personas. Table~\ref{tab:persona_instruction_examples} presents several representative instruction examples corresponding to different personas.

Taking the \textbf{Courier / Delivery worker} persona as an example, the instructions are clearly oriented toward task execution and road navigation contexts. 
The language heavily relies on traffic-related facilities and drivable elements as core references, such as \emph{intersection}, \emph{multi-lane road}, \emph{lane markings}, and \emph{parking area}. 
Path descriptions are closely organized around intersection choices, lane following, and parking area localization, reflecting a strong preference for efficiency and accessibility.

Instructions from the \textbf{Teacher / Educator} persona exhibit a distinct instructional and guidance-oriented characteristic. 
The expressions are typically organized in a progressive manner, with an emphasis on maintaining directional consistency and intermediate states (e.g., \emph{keep your heading}, \emph{maintain the same heading}) to reduce cognitive load during understanding and execution.
In addition, landmarks are described in a more detailed and reproducible manner, such as explicit specifications of track color or lane line styles, ensuring that each step is grounded in stable and clear perceptual cues.

In contrast, instructions from the \textbf{Retired adult} persona place greater emphasis on communication comfort and everyday experience. 
The language style is more conversational, with a gentle and reassuring tone (e.g., \emph{Alright, nice and easy now}, \emph{When you're ready}, \emph{go forward a bit more}). 
These instructions tend to rely on landmarks closely related to daily life, such as \emph{footbridge}, \emph{creek}, and \emph{sheds}. 
Precise constraints are intentionally reduced, with greater focus on situational guidance and pace control.

Overall, different personas exhibit systematic differences in information focus, path organization, and interaction style. 
These differences are closely associated with their respective occupational backgrounds and life experiences, enabling AirNav to generate navigation instructions that better align with real human expression habits and demonstrate higher instruction diversity.

\begin{table*}[t]
\centering
\small
\setlength{\tabcolsep}{6pt}
\renewcommand{\arraystretch}{1.15}
\begin{tabular}{p{3.5cm} p{12cm}}
\hline
\textbf{Persona} & \textbf{Instruction Example} \\
\hline
Courier / Delivery Worker &
Turn left and move forward until the curved road intersection with white painted lines and a central grass island is ahead. At the intersection, turn left and move forward. Move forward, then turn right and move forward until you reach the curved roadway intersection with grassy medians and a large tree in the middle. Move forward until the multi-lane road with white lane markings and a pedestrian crossing appears beside parked cars and industrial buildings. Move forward along this road, then turn left toward the small parking area by the light-gray industrial buildings. Move forward to the bright red car and stop. \\
\hline
Teacher / Educator &
Begin by turning right until the tan running track with a solid white lane line near the infield edge is aligned ahead. Move forward toward it, then make a slight right and continue forward along the track edge until you reach the section with the white lane line. From this point, keep your heading and proceed straight toward the curved arc of blue-and-white seating outside the track, continuing forward until you arrive beside the seating. Maintain the same heading and advance straight toward the row of tightly parked white cars along the side of the long gray-roof workshop, continuing until you reach that row. Continue straight along the workshop side, then make a slight left and move forward to the small red car parked in the lot beside the gray-roof warehouses, stop. \\
\hline
Retired Adult &
Alright, nice and easy now. Begin by turning right, then move forward at a comfortable pace. When you’re ready, turn right again and continue moving forward until you come to the large industrial building with several gray metal roofs and a parking lot that has white trucks beside it. From there, move forward a touch and turn left, then keep moving forward, slow and steady, toward the row of white and blue trucks in the lot. Continue moving forward past those trucks, and when you feel settled, turn left and go forward until the short white footbridge over the narrow dark creek is ahead. At the footbridge, turn right and keep moving forward along the yard until you reach the row of narrow gray‑roofed sheds with multiple bays. Go forward a bit more, then turn left and move forward toward the blue car parked among other vehicles in the lot, and stop. \\
\hline
\end{tabular}
\caption{Representative Instruction Examples for Different Personas}
\label{tab:persona_instruction_examples}
\end{table*}

\section{Instruction Naturalness Evaluation Prompt}
\label{app:prompt_naturalness_evaluation}

\begin{prompt}{Prompt}
\textbf{\small Role:}

You are an expert language evaluation assistant for UAV navigation instructions.
\\
\\    
\textbf{\small Objective:}

Your task is to assess the Naturalness of a navigation instruction as it would be spoken by a real human guiding a UAV in a real-world scenario.
\\
\\
\textbf{\small Evaluation Criteria:}

Evaluate the instruction based on the following criteria:

1. Naturalness: Whether the instruction sounds like spontaneous human speech rather than a rigid, scripted, or templated command.

2. Practicality: Whether the instruction provides actionable route guidance through landmarks, relative directions, or intermediate cues, rather than low-level action enumeration.

3. Human Alignment: Whether the wording and structure align with how a human would naturally phrase a navigation request in everyday use.
\\
\\
\textbf{\small Rating Scale:}

Rate it on a scale from 1 to 5:

- 1 = very unnatural (robotic, templated, or action-list-like)

- 2 = somewhat unnatural (syntactically valid but awkward or artificial)

- 3 = neutral (reasonable but not strongly human-like)

- 4 = mostly natural (sounds like something a person might naturally say)

- 5 = very natural (fluent, realistic, and clearly human-like)
\\
\\
\textbf{\small Output Requirement:}

Your output must be a single integer from 1 to 5 representing the overall naturalness score. Only output the number, and do not provide any explanation.
\\
\\
\textbf{\small Task:}

Evaluate the following UAV navigation instruction according to the criteria above.
\\
\\
\textbf{\small Navigation Instruction:}

\verb|{navigation instruction}|
\end{prompt}

\paragraph{Evaluation Protocol.}
For each dataset, we randomly sample 2,000 instructions and score them using the prompt above with GPT-4o.
To reduce randomness from single-pass evaluation, each instruction is independently scored three times, and the final score is the average of the three results.
Note that datasets whose instructions only describe the target object without any intermediate path or process-level guidance are excluded from the comparison to avoid unfair evaluation.

\section{Human Annotation Setup and Statistical Analysis}
\label{app:human_annotation}

To validate the reliability of LLM-based automatic evaluation for instruction naturalness, we introduce human annotation as a calibration reference and analyze the consistency between LLM scores and human judgments.

\paragraph{Human Annotation Setup}
We randomly sample a total of 500 instructions from all datasets involved in the instruction naturalness analysis, with approximately 100 instructions drawn from each dataset, to form the human-annotated subset. Each instruction is independently evaluated by three annotators. All annotators are fluent in English and have a basic understanding of UAV VLN tasks. Human annotation follows the same scoring scheme as the automatic evaluation. Prior to annotation, all annotators are provided with a unified annotation guideline to ensure a consistent understanding of the evaluation dimensions and scoring scale.

\paragraph{Inter-Annotator Agreement}
We first analyze the inter-annotator agreement to assess the stability of human judgments. Specifically, we adopt Krippendorff’s $\alpha$ as the agreement metric. The results show that Krippendorff’s $\alpha$ reaches 0.70, indicating moderate to substantial agreement among annotators and a stable consensus in their interpretation of the evaluation criteria.

\paragraph{Human--LLM Correlation}
We further examine the correlation between human annotations and LLM-based scores. For each instruction, the scores from the three annotators are averaged to obtain a human reference score, which is then compared with the corresponding LLM score using Spearman’s rank correlation coefficient ($\rho$). The results show a strong positive correlation, with Spearman’s $\rho$ reaching 0.74, suggesting that LLM-based scoring effectively captures the overall trend of human judgment on instruction naturalness.

\paragraph{Discussion}
The above results indicate that LLM-based automatic evaluation is highly consistent with human judgments in instruction naturalness assessment, while significantly reducing evaluation costs. Therefore, we adopt LLM-based scoring as the instruction naturalness evaluation tool in the main experiments to enable systematic comparison across large-scale datasets.

\section{Case Study on Instruction Naturalness}
\label{app:case_study_instruction_naturalness}

We further conduct a qualitative analysis of instruction naturalness across different datasets through case studies. 
Specifically, we randomly select two instruction examples from each dataset, as shown in Table~\ref{tab:case_study_instruction_naturalness}, and compare them along multiple dimensions related to naturalness.

Taking AerialVLN as an example, its instructions are typically composed of a sequence of low-level actions, with repeated use of expressions such as \emph{turn around}, \emph{continue straight}, and \emph{go over the buildings}. 
Although macro-level landmarks such as lakes and fountains are mentioned, these landmarks are not further specified with perceptually distinguishable attributes. 
In addition, turning descriptions such as \emph{turn up and left} are semantically vague and difficult to precisely associate with concrete spatial changes. 
This style is closer to programmatic control commands than to natural language instructions used by humans during real navigation.

Instructions in the LANI dataset exhibit clearer executability, as they often constrain motion trajectories using precise turning angles (e.g., \emph{Turn 60 degrees}), thereby reducing ambiguity.
However, this highly numerical form of expression does not align well with common human navigation communication habits, as people rarely rely on frequent and exact angle specifications in real-world scenarios. 
Moreover, the referenced landmarks are typically simple and local small objects (e.g., \emph{green cactus} and \emph{potted plant}), whose visual salience and recognizability are limited from a UAV perspective, weakening the alignment between linguistic descriptions and visual perception.

In contrast, instructions in our dataset demonstrate characteristics that are closer to human navigation language. 
They tend to rely on landmarks with higher visual salience and discriminability, which are further specified through attributes such as color, shape, and spatial relationships (e.g., \emph{dark-gray rectangular roof}, \emph{curved dirt racetrack}, and \emph{single white car}), thereby reducing potential ambiguity. 
In terms of action description, our instructions more frequently adopt progressive and relative expressions (e.g., \emph{make a slight left} and \emph{continue past \ldots, then turn \ldots}), rather than relying heavily on precise angle values, making them more consistent with human navigation language usage.

\begin{table*}[t]
\centering
\small
\setlength{\tabcolsep}{6pt}
\begin{tabular}{p{2.5cm} p{6.2cm} p{6.2cm}}
\hline
\textbf{Dataset} & \textbf{Example 1} & \textbf{Example 2} \\
\hline
LANI &
Move forward and stay to the right of the red ball. Continue traveling in a straight line until you pass the brown chair, which will be on your left. Just after passing the chair, turn almost 90 degrees to the left, and continue in a straight line until you pass a green cactus on your left. Turn 60 degrees to your left, and continue in a straight line until you pass a potted plant on your left. Turn 45 degrees left just past the potted plant, and then continue in a straight line until a traffic cone is immediately on your right. &
Move forward towards the lamp and move past it on its left. Move forward towards the chest and move past it on its left. Stop before hitting the red fence and turn right. Move forward towards the grey object and move past it on its left. Stop before hitting the white fence. \\
\hline
AVDN &
"[INS] Destination is a long row of short cargo containers bisecting the island of dirt at your eight o'clock direction..",
"[QUE] I am on top of the many small containers, Can I see the destination? Which direction should I go?."
"[INS] Northeast, fly to your right and turn around and go the upposit direction  just a few feet to your destination."
"[QUE] I move to the northeast, Am I near the destination? Which direction should I go?."
"[INS] Turn 180 degrees.  Go straight until you are over a white container.  That is your destination.". &
"[INS] Please go to the southeast direction at 5 o'clock. The destination is blue containers."
"[QUE] I am on top of many containers, Am I near the destination? Where should I go now?"
"[INS] Yes, you are very close to your destination. Turn to your 5 o'clock and go straight forward from there and you will be right on top of your destination." \\
\hline
AerialVLN &
Go up the building and fly to the left near the lake. stop at the middle of the lake and turn up and left and prepare for take off. go over the trees and bushes until you see the fountain. go over the fountain and over one building until you see the edge of the ocean. turn around and head towards the spotted lake and land near the trees. turn around and continue straight and over the red tree. turn around and continue straight. &
Take off and turn right and move forward and cross the buildings terrace and turn left. now move forward and go over the buildings towards the white building terrace and turn left. now move forward cross the buildings and go over the grass and roads and reach the under the building. now move forward and turn right and go over the buildings and towards the brown building terrace and stay there. \\
\hline
OpenFly &
Advance forward to a multi-colored residential area predominantly featuring beige and yellow tones. The area consists of a cluster of low-rise residential buildings with flat rooftops and tree - lined parallel streets. Slightly turn left and proceed straight to reach a yellow medium-sized residential building known for its rectangular structure, repetitive windows, and dark roofs. Shift right to find a medium-sized beige and gray residential building with multiple floors and uniform windows. Move ahead to encounter a medium-sized building with a brown and gray exterior, rooftop garden, and a rectangular shape, surrounded by structures of comparable scale. Slightly turn left again and proceed straight to it. &
Proceed to the gray skyscraper building, then turn right towards the green golf course with trees and a pond, an expansive scenic outdoor area. Continue straight ahead, subtly veer left, and proceed straight towards the large green tree with coniferous leaves. \\
\hline
Ours &
Turn left to face the grassy park with scattered trees and a curved edge path. Move forward, make a slight left, and reach this park. Continue straight to the next grassy area with scattered trees and a curved path beside the residential streets. Keep straight, then turn right toward the two‑lane road with houses on both sides and move to the road. Proceed forward along the road, then turn left toward the dark‑gray rectangular roof of the terraced house and continue to it. Continue ahead toward the parking lot, turn right to the black car parked in a row, move to the car, stop. &
Move straight ahead toward the wide multi‑lane road with white dashed markings beside the unpaved lot, continuing forward until you reach it. Stay on this line, then turn right and proceed straight toward the curved dirt racetrack surrounding the grassy field. Continue past the racetrack, then turn left and head straight toward the narrow waterway bordered by trees. Keep forward, then turn right and continue straight to the large industrial building with a gray roof and adjacent parking lot. From this building, turn left and move straight along the main road until the single white car in the traffic lane is ahead; approach it and stop. \\
\hline
\end{tabular}
\caption{Instruction Examples from Different Datasets}
\label{tab:case_study_instruction_naturalness}
\end{table*}

\section{Prompt Template for AirVLN-R1}
\label{app:prompt}

\begin{prompt}{Prompt}
\textbf{\small Role:}

You are an expert navigation assistant for a UAV (Unmanned Aerial Vehicle) flight simulator.
\\
\\    
\textbf{\small Task Objective:}

The UAV operates in an urban environment with visible roads, buildings, and landmarks.
Your task is to predict the next sequence of UAV actions based on:

1. A given natural language navigation instruction,

2. The current state of the UAV, including its position and heading angle,

3. The current first-person UAV view image,

4. Up to four historical first-person view images from previous time steps (if available),

5. The previously executed UAV actions (if available).
\\
\\
\textbf{\small Text Input:}

- Navigation instruction: \verb|{Instruction}| \\
- Current state of the UAV: \verb|{Current State}| \\
- Previously executed actions: \verb|{Historical Action sequence}| \\
  (A list of past actions the UAV has taken, in chronological order.)
\\
\\
\textbf{\small Image Input:}

UAV (Unmanned Aerial Vehicle) View Sequence

- Historical views (from oldest to newest) show the UAV’s past observations.\\ 
- The last image represents the UAV’s current view.  \\
- In all images, the top of the frame corresponds to the UAV’s forward direction (its heading).
\\
\\
\textbf{\small Step-by-Step Action Planning:}

Based on the navigation instruction, the UAV’s current state, the previously executed actions (which can help infer the UAV’s current orientation and progress), and the provided images, predict how the UAV should move step by step to follow the instruction accurately.
\\
\\
\textbf{\small Prediction Rules:}

1. Predict no more than 8 future actions for the UAV to execute.

2. If the target location is reachable in fewer than 8 actions, output less than 8 actions sequence and end with "STOP". Otherwise, it clearly requires more than 8 actions to approach the target, output exactly 8 future actions.

3. You must output "STOP" if the UAV has already reached the described target.

4. Output a JSON list of actions, in the exact order they should be executed.

5. Do not include any explanations, reasoning, or additional text — only output the JSON list.
\\
\\
\textbf{\small Discrete Action Space:}

    - MOVE\_FORWARD: move straight 5 meters in the current heading \\
    - TURN\_LEFT: rotate left 30 degrees \\
    - TURN\_RIGHT: rotate right 30 degrees \\
    - STOP: stop the flight
\\
\\
\textbf{\small Output Format Examples:}

["TURN\_RIGHT", "TURN\_RIGHT", "MOVE\_FORWARD", "MOVE\_FORWARD", "MOVE\_FORWARD", "MOVE\_FORWARD", "MOVE\_FORWARD", "MOVE\_FORWARD"]

["MOVE\_FORWARD", "MOVE\_FORWARD", "STOP"]

["STOP"]
\end{prompt}

\section{Baseline Model Descriptions}
\label{app:baselines}
We provide a brief overview of the baseline models evaluated on the AirNav benchmark.
\begin{itemize}[leftmargin=*, itemsep=0pt]
    \item \textbf{Seq2Seq} \cite{anderson2018vision}: A classic end-to-end model that directly encodes both language and visual inputs, mapping them to corresponding action sequences. This model serves as the basic comparison method.
    \item \textbf{CMA} \cite{hu2024generating}: A model that employs a cross-modal attention mechanism, effectively integrating language and visual information to make navigation decisions.
    \item \textbf{Qwen2.5-VL (7B / 32B)} \cite{bai2025qwen25vltechnicalreport}: A MLLM developed and open-sourced by Qwen team, Alibaba Cloud. It supports both image input and language understanding, with different parameter sizes, making it suitable for analyzing the impact of model capacity on navigation tasks.
    \item \textbf{Qwen3-VL-235B-A22B} \cite{bai2025qwen3vltechnicalreport}: A large-scale Mixture-of-Experts vision-language model released by the Qwen team, with 235B total parameters and 22B activated parameters per token, making it a strong reference model for evaluating the upper-bound performance of general-purpose MLLMs in navigation tasks.    
    \item \textbf{LLaMA-3.2-11B-Vision} \cite{grattafiori2024llama}: An open-source MLLM released by Meta, which combines image recognition and language reasoning abilities to evaluate the perceptual and decision-making capabilities of LLM in navigation tasks.
    \item \textbf{GPT-4o} \cite{openai2024gpt4ocard}: A powerful multimodal general-purpose model launched by OpenAI, supporting joint text and image input. With its excellent reasoning capabilities, it serves as an important reference for evaluating task performance.

    \item \textbf{$\pi_0$} \cite{black2024pi_0}: A generalist Vision-Language-Action model pre-trained on large-scale robot manipulation data, built upon PaliGemma with a flow-matching action expert. Since $\pi_0$ originally outputs continuous actions, we replace its action head with a discrete action prediction head to adapt it to the UAV VLN task.

    \item \textbf{OpenVLA} \cite{kim2025openvla}: An open-source 7B-parameter Vision-Language-Action model built upon the Prismatic-7B VLM backbone (Llama 2 7B with fused DINOv2 and SigLIP visual encoders), fine-tuned on 970K real-world robot manipulation trajectories from the Open X-Embodiment dataset.

    \item \textbf{Uni-NaVid} \cite{zhang2024uni}: A unified video-based VLA model trained on large-scale multi-task navigation data, covering diverse embodied navigation settings including object navigation, instruction following, and person tracking.

    \item \textbf{OpenFly-Agent} \cite{2025OpenFly}: A keyframe-aware VLN model specifically designed for aerial navigation, which leverages key visual observations to improve navigation efficiency and accuracy.
\end{itemize}

    

\section{Hyperparameter Settings for SFT and RFT Stages}
\label{app:hyperparameter_settings}

The hyperparameter settings used in the SFT and RFT stages are summarized in Table~\ref{tab:train-config}.

\begin{table*}[t]
\centering
\tiny
\setlength{\tabcolsep}{3pt}
\resizebox{\textwidth}{!}{
\begin{tabular}{
>{\centering\arraybackslash}p{1.0cm} |
>{\centering\arraybackslash}p{1.8cm}
>{\centering\arraybackslash}p{0.9cm}
>{\centering\arraybackslash}p{1.0cm}
>{\centering\arraybackslash}p{0.8cm}
>{\centering\arraybackslash}p{0.9cm}
>{\centering\arraybackslash}p{1.6cm}
>{\centering\arraybackslash}p{1.3cm}
>{\centering\arraybackslash}p{1.2cm}
>{\centering\arraybackslash}p{1.5cm}
>{\centering\arraybackslash}p{0.6cm}
>{\centering\arraybackslash}p{0.6cm}
>{\centering\arraybackslash}p{0.6cm}
>{\centering\arraybackslash}p{0.6cm}
>{\centering\arraybackslash}p{0.6cm}
}
\toprule
\textbf{Stage} &
\textbf{Framework} &
\textbf{Batch} &
\textbf{LR} &
\textbf{Epochs} &
\textbf{Steps} &
\textbf{\#Historical Views} &
\textbf{Group Size} &
\textbf{KL Coeff} &
\textbf{Sampling Temp} &
$\boldsymbol{\lambda_1}$ &
$\boldsymbol{\lambda_2}$ &
$\boldsymbol{\alpha}$ &
$\boldsymbol{\beta}$ &
$\boldsymbol{\gamma}$ \\
\midrule
SFT &
LLaMA-Factory &
80 &
$1\times10^{-4}$ &
2 &
-- &
4 &
-- &
-- &
-- &
-- &
-- &
-- &
-- &
-- \\
RFT &
verl &
96 &
$1\times10^{-6}$ &
-- &
1500 &
4 &
5 &
0.001 &
1.0 &
1 &
1 &
1 &
0.1 &
0.1 \\
\bottomrule
\end{tabular}
}
\caption{Hyperparameter settings for the SFT and RFT training stages.}
\label{tab:train-config}
\end{table*}



\section{Real-World Test Setup and Test Cases}
\label{app:real_world_setup_and_cases}

\begin{figure*}[t]
  \centering
  \includegraphics[width=0.97\textwidth]{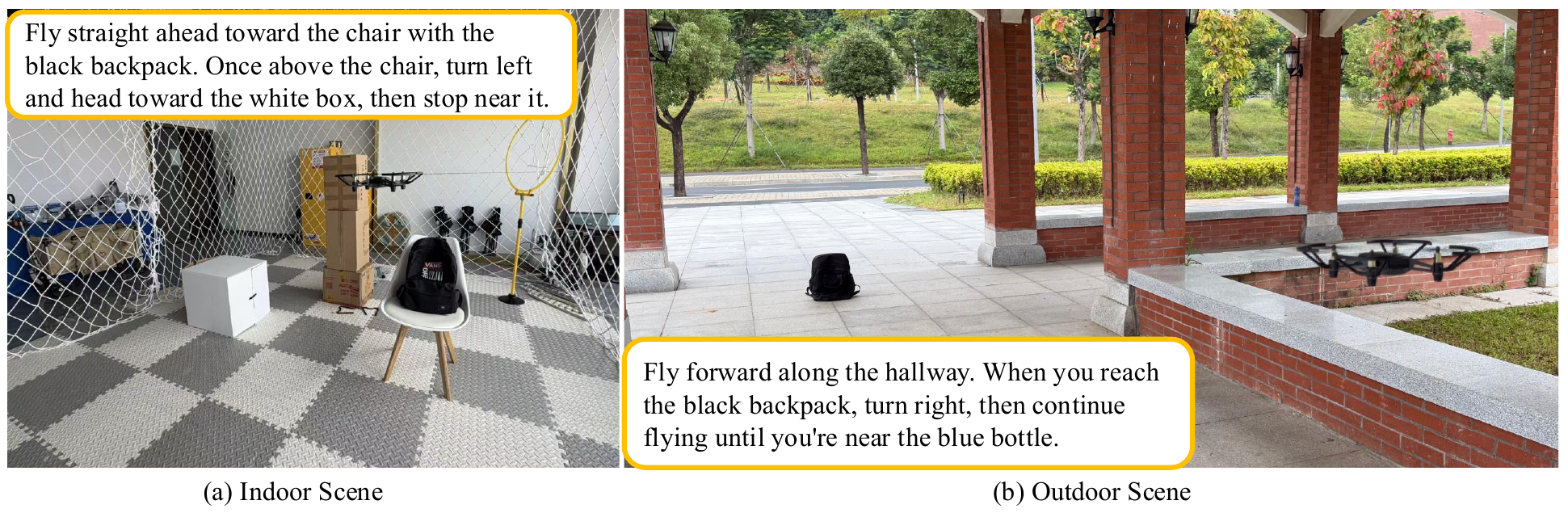} 
  \vspace{-8pt}
  \caption{Real-World UAV VLN Deployment in Indoor and Outdoor Scenes.}
  \label{fig:real_world_case}
\end{figure*}

The experimental UAV was a DJI Tello TLW004, equipped with a camera, a Vision Positioning System, a barometer, an infrared sensor, and an inertial measurement unit.
During testing, the UAV first transmitted its real-time images and state information to a local laptop via a network. The laptop then forwarded the data to a remote server or cloud platform where the model was deployed. Upon receiving all input information, the model generated the corresponding action sequence, which was subsequently executed by the UAV through its built-in motion control API.
In both indoor and outdoor settings, the UAV maintains a fixed altitude throughout each navigation task; ascending and descending actions are excluded from the action space. Navigation trajectories are therefore planar, which is consistent with the benchmark design and simplifies control for fair cross-method comparison.
Fig.~\ref{fig:real_world_case} shows two representative real-world navigation tasks conducted in indoor and outdoor environments.

\section{Persona-Level Evaluation Results}
\label{app:persona_results}

Table~\ref{tab:persona_results} reports the SR (\%) of all models on the test-unseen split under each of the 10 user personas.

\begin{table*}[t]
\centering
\tiny
\setlength{\tabcolsep}{2.5pt}
\resizebox{\textwidth}{!}{
\begin{tabular}{l | *{10}{>{\centering\arraybackslash}p{0.8cm}}}
\toprule
\textbf{Method} & \textbf{P1} & \textbf{P2} & \textbf{P3} & \textbf{P4} & \textbf{P5} & \textbf{P6} & \textbf{P7} & \textbf{P8} & \textbf{P9} & \textbf{P10} \\
\midrule
\multicolumn{11}{l}{\textit{P1: Primary school student \quad P2: University student \quad P3: Office worker \quad P4: Community staff \quad P5: Courier/Delivery worker}} \\
\multicolumn{11}{l}{\textit{P6: Urban tourist \quad P7: Retired adult \quad P8: Teacher/Educator \quad P9: Community volunteer \quad P10: Advanced navigation user}} \\
\midrule


Random              & 0.65 & 0.85 & 1.05 & 0.69 & 0.75 & 0.52 & 0.64 & 0.91 & 0.72 & 0.96 \\
Seq2Seq             & 1.19 & 1.76 & 1.05 & 1.43 & 1.40 & 0.92 & 0.93 & 1.40 & 1.60 & 1.47 \\
CMA                 & 4.46 & 4.73 & 4.32 & 4.47 & 4.62 & 3.84 & 3.50 & 6.14 & 5.26 & 4.93 \\
Qwen2.5-VL-7B       & 1.55 & 1.88 & 1.69 & 1.83 & 1.40 & 2.01 & 1.14 & 2.13 & 2.16 & 0.96 \\
Qwen2.5-VL-32B      & 2.14 & 3.34 & 2.85 & 3.55 & 2.42 & 3.44 & 2.72 & 3.34 & 3.54 & 2.43 \\
Qwen3-VL-235B-A22B  & 4.34 & 4.79 & 5.01 & 5.56 & 5.00 & 4.99 & 4.00 & 7.84 & 6.64 & 3.39 \\
LLaMA-3.2-11B-Vision & 2.27 & 2.44 & 2.22 & 0.00 & 0.00 & 1.01 & 1.82 & 1.30 & 1.22 & 1.47 \\
GPT-4o                    & 4.16 & 4.00 & 4.00 & 5.33 & 4.46 & 4.76 & 3.57 & 5.78 & 4.76 & 4.71 \\
\midrule
$\pi_0$ & 4.76 & 6.00 & 4.64 & 5.21 & 6.56 & 5.05 & 3.57 & 5.65 & 4.81 & 5.80 \\
Uni-NaVid             & 13.66 & 16.92 & 14.01 & 15.64 & 16.87 & 13.47 & 11.34 & 17.26 & 17.21 & 16.04 \\

Qwen2.5-VL-7B SFT-only  & 33.79 & 43.30 & 39.09 & 39.26 & 41.16 & 38.53 & 38.10 & 44.50 & 39.26 & 43.23 \\
Qwen2.5-VL-7B RFT-only  & 2.08  & 2.18  & 2.11  & 2.75  & 2.63  & 2.52  & 1.50  & 3.22  & 2.66  & 2.21  \\
\textbf{AirVLN-R1 (Ours)} & \textbf{44.74} & \textbf{55.67} & \textbf{52.27} & \textbf{50.60} & \textbf{53.79} & \textbf{48.97} & \textbf{45.82} & \textbf{57.57} & \textbf{52.08} & \textbf{56.70} \\


\bottomrule
\end{tabular}
}
\caption{SR (\%) of all models under each user persona on the test-unseen split. 
}
\label{tab:persona_results}
\end{table*}

\section{Additional Ablation Studies}
\label{app:additional_ablation}

\subsection{Historical View Image Selection}
\label{app:ablation_history}

To systematically analyze the impact of different historical view image selection strategies on navigation performance, we compare four strategies under the SFT+RFT training paradigm and evaluate them on the test-unseen split.
Specifically, the evaluated strategies include:
\begin{itemize}
    \item \textbf{No-History}, which uses only the current view image without incorporating any historical observations;
    \item \textbf{Last-K}, which always selects the most recent $K$ historical view images;
    \item \textbf{Uniform-K}, which uniformly samples $K$ images within a fixed historical window;
    \item \textbf{Progressive Interval Sampling (Ours)}, which adopts a non-uniform sampling strategy with progressively increasing temporal intervals.
\end{itemize}

Table~\ref{tab:history_image_ablation} reports the performance of different strategies on the test-unseen split. 
Several observations can be drawn.
First, the \textbf{No-History} baseline, which relies solely on the current view image, performs the worst, with NE as high as 121.6 and SR of only 22.96\%.
This result indicates that single-frame visual observations are insufficient to provide adequate temporal and spatial context for reliable decision-making.
Second, the \textbf{Last-K} strategy significantly improves performance by retaining the most recent $K$ historical images, enabling the model to capture short-term motion trends and local environmental changes.
As a result, the SR increases to 45.1\%.
However, since this strategy ignores earlier but potentially important observations, the model’s ability to reason about long-term navigation progress remains limited.
Third, \textbf{Uniform-K} further improves performance by uniformly sampling historical images within a fixed window, allowing the model to access observations at multiple temporal scales.
This leads to a lower NE of 41.1 and a higher SR of 49.83\%, suggesting that covering a longer temporal span of historical information helps the model better understand the global path structure.
Nevertheless, due to its fixed sampling interval, this strategy lacks an explicit mechanism to model the temporal importance difference between recent and distant observations.
Finally, \textbf{Progressive Interval Sampling} achieves the best performance across all metrics, yielding the lowest NE (39.8) and the highest SR (51.82\%) on the test-unseen split. 
Compared to Uniform-K, it further improves both navigation success and path execution efficiency. 
These results demonstrate that non-uniform sampling with progressively increasing intervals more effectively balances fine-grained short-term perception and long-term contextual modeling.

\begin{table*}[t]
\centering
\tiny
\setlength{\tabcolsep}{4pt}
\begin{tabular}{
>{\centering\arraybackslash}p{4.2cm} |
>{\centering\arraybackslash}p{1.0cm}
>{\centering\arraybackslash}p{1.0cm}
>{\centering\arraybackslash}p{1.0cm}
>{\centering\arraybackslash}p{1.0cm}
}
\toprule
\textbf{Method} & NE$\downarrow$ & SR$\uparrow$ & OSR$\uparrow$ & SPL$\uparrow$ \\
\midrule

No-History & 121.6 & 22.96 & 34.79 & 21.84 \\
Last-K & 56.8 & 45.1 & 56.15 & 43.96 \\
Uniform-K & 41.1 & 49.83 & 60.14 & 48.72 \\
Progressive Interval Sampling (Ours) & \textbf{39.8} & \textbf{51.82} & \textbf{62.73} & \textbf{50.66} \\

\bottomrule
\end{tabular}
\caption{Ablation Study of Historical View Image Selection Strategies (test-unseen)}
\label{tab:history_image_ablation}
\end{table*}

\subsection{Ablation on Reward Components}
\label{app:ablation_reward}

To further quantify the contribution of each reward component in the RFT stage, we conduct a series of ablation experiments based on the same SFT-initialized model. Specifically, different components of the reward function are removed, and all models are evaluated on the test-unseen split for comparison.
The following ablation settings are considered:
\begin{itemize}
    \item \textbf{w/o Subgoal State Alignment}: removing the Distance-to-Subgoal and Heading Angle Alignment reward, while retaining the Stop Consistency and Format Reward;
    \item \textbf{w/o Stop Consistency}: removing the Stop Consistency reward;
    \item \textbf{w/o Format Reward}: removing the reward that enforces output validity.
\end{itemize}

Table~\ref{tab:reward_ablation} reports the experimental results under different settings on the test-unseen split, where \textbf{SFT-only} serves as the baseline without RFT. The results lead to the following observations.
First, the Subgoal State Alignment reward is the primary driver of performance improvement. Removing this component results in the most significant degradation across all metrics, with NE increasing from 39.8 to 46.1 and SR dropping sharply from 51.82\% to 46.00\%. This indicates that Subgoal State Alignment plays a critical role in guiding effective path planning.
Second, the Stop Consistency reward affects the reliability of \textbf{stop} decisions.
Without this reward, the model exhibits a substantially higher frequency of both early-stop and missed-stop, leading to a noticeable decline in SR (51.82\% $\rightarrow$ 47.65\%). This result highlights the importance of explicitly supervising \textbf{when to stop} in UAV VLN tasks for stable execution and successful task completion.
Third, the Format reward has a relatively limited impact on the final performance metrics but contributes positively to training stability and model usability. 
Since the output structure is already well constrained during the SFT stage, removing the Format reward leads only to a slight performance drop. 
Nevertheless, this reward helps reduce invalid generations, thereby improving the stability of the RFT and the reliability of practical deployment.
Overall, the three reward components exhibit complementary roles: the Subgoal State Alignment reward ensures that the agent \textbf{moves in the correct direction}, the Stop Consistency reward encourages the agent to \textbf{stop at the appropriate time}, and the Format reward guarantees that the \textbf{outputs are valid}. 
Together, they form a systematic reward design tailored for UAV VLN, resulting in improved performance and stronger generalization.

\begin{table*}[t]
\centering
\tiny
\setlength{\tabcolsep}{4pt}
\begin{tabular}{
>{\centering\arraybackslash}p{4.2cm} |
>{\centering\arraybackslash}p{1.0cm}
>{\centering\arraybackslash}p{1.0cm}
>{\centering\arraybackslash}p{1.0cm}
>{\centering\arraybackslash}p{1.0cm}
}
\toprule
\textbf{Method} & NE$\downarrow$ & SR$\uparrow$ & OSR$\uparrow$ & SPL$\uparrow$ \\
\midrule
Qwen2.5-VL-7B SFT-only & 47.5  & 40.20 & 53.21 & 39.15 \\
SFT + RFT (Full Reward, AirVLN-R1) & 39.8 & 51.82 & 62.73 & 50.66 \\
SFT + RFT (w/o Subgoal State Alignment) & 46.1 & 46.00 & 60.58 & 44.90 \\
SFT + RFT (w/o Stop Consistency) & 44.2 & 47.65 & 61.02 & 46.53 \\
SFT + RFT (w/o Format Reward) &  40.7 & 51.22 & 62.08 & 50.06 \\
\bottomrule
\end{tabular}
\caption{Ablation results of reward components in the RFT Stage (test-unseen)}
\label{tab:reward_ablation}
\end{table*}

\subsection{Sensitivity Analysis of Reward Weights}
\label{app:ablation_sensitivity}

The overall reward function involves five weighting parameters: $\lambda_1$, $\lambda_2$, $\alpha$, $\beta$, and $\gamma$.
The component-level ablation in Table~\ref{tab:reward_ablation} has already demonstrated that Subgoal State Alignment (governed by $\lambda_1$ and $\lambda_2$) is the dominant contributor to performance, while Stop Consistency ($\alpha$) and Format Reward ($\beta$, $\gamma$) have comparatively smaller and stable effects.
We therefore focus the sensitivity analysis on $\lambda_1$ and $\lambda_2$, as their relative weighting directly controls the balance between distance alignment and heading alignment within the most critical reward component.

Keeping all other RFT settings strictly unchanged (including the initialized model, GRPO configuration, learning rate, training steps, and the stop/format rewards), we conduct a $3 \times 3$ grid search by varying $\lambda_1$ and $\lambda_2$ over $\{0.5, 1.0, 2.0\}$, and evaluate SR on test-unseen.

\begin{table}[h]
\centering
\small
\begin{tabular}{c|ccc}
\toprule
\diagbox{$\lambda_2$}{$\lambda_1$} & 0.5 & 1.0 & 2.0 \\
\midrule

0.5 & 50.83 & 51.46 & 51.32 \\
1.0 & 51.20 & \textbf{51.82} & 51.41 \\
2.0 & 50.42 & 50.95 & 50.63 \\

\bottomrule
\end{tabular}
\caption{Sensitivity analysis of $\lambda_1$ and $\lambda_2$ (SR\% on test-unseen).}
\label{tab:sensitivity_lambda}
\end{table}

As shown in Table~\ref{tab:sensitivity_lambda}, performance remains stable across the tested range, with the maximum SR fluctuation within 1.5 percentage points.
The default setting ($\lambda_1=1$, $\lambda_2=1$) achieves slightly better results, but multiple neighboring configurations yield comparable performance, indicating that the method does not depend on a precise weight choice.

\section{Real-World Evaluation Results}
\label{app:real_world_results}

Table~\ref{tab:real_world_result} summarizes the performance, resource consumption, 
and inference latency of different methods in real-world UAV VLN experiments.
\begin{table*}[t]
\centering
\tiny
\setlength{\tabcolsep}{3pt}
\begin{tabular}{>{\centering\arraybackslash}p{4.2cm} | >{\centering\arraybackslash}p{1.2cm} >{\centering\arraybackslash}p{1.2cm} >{\centering\arraybackslash}p{2.2cm} >{\centering\arraybackslash}p{2.2cm} >{\centering\arraybackslash}p{2.2cm}}
\toprule
\textbf{Method} & \textbf{NE$\downarrow$} & \textbf{SR$\uparrow$} & \textbf{Test Device} & \textbf{GPU Memory Usage (GB)} & \textbf{Inference Latency (s/step)} \\
\midrule

Seq2Seq & N/A & 1/200 & RTX 4090 * 1 & 0.22 & 0.080 \\
CMA & N/A & 2/200 & RTX 4090 * 1 & 0.27 & 0.075 \\
Qwen2.5-VL-7B & 108.4 & 8/200 & RTX 4090 * 1 & 20.40 & 0.840 \\
Qwen2.5-VL-32B & 88.2 & 17/200 & A100 80GB * 1 & 67.64 & 2.511 \\
GPT-4o & 73.1 & 34/200 & Cloud & / & 3.674 \\
AirVLN-R1 (Ours) & 70.6 & 53/200 & RTX 4090 * 1 & 20.40 & 0.854 \\
\bottomrule
\end{tabular}

\vspace{0.5em}
\scriptsize
\textbf{Note.} Seq2Seq and CMA fail to generate valid trajectories in most real-world episodes; their NE values are therefore not statistically meaningful and reported as N/A.
\caption{Comparison of performance, resource usage, and inference latency on real-world test.}
\label{tab:real_world_result}
\end{table*}






\section{Analysis of Resource Cost and Inference Efficiency}
\label{app:resource_cost}

Table~\ref{tab:real_world_result} shows the performance and resource usage comparison of different models. The analysis is as follows:
\begin{itemize}[leftmargin=*, itemsep=0pt]
    \item \textbf{Traditional models are lightweight but impractical:} Seq2Seq and CMA exhibit low inference latency and memory consumption; however, they nearly fail to accomplish any navigation tasks in real-world tests, succeeding on only 1--2 out of 200 tasks, resulting in limited practical usability.
    \item \textbf{Deployment cost grows substantially for large models:} Although Qwen2.5-VL-32B achieves better performance than Qwen2.5-VL-7B, it incurs significantly higher computational demands and inference latency, requiring an A100 80GB GPU, which makes it difficult to meet the real-time requirements.
    \item \textbf{Cloud models introduce controllability and real-time risks:} GPT-4o avoids local memory usage via cloud inference, but still has a latency of 3.674~s/step and may impose potential constraints on system controllability and data privacy.
    \item \textbf{AirVLN-R1 balances performance and efficiency:} AirVLN-R1 achieves the highest SR in real-world tests while keeping the computational overhead acceptable, providing a more balanced trade-off between performance and deployment cost.
\end{itemize}

\end{document}